\documentclass[11pt]{article}

\usepackage[preprint]{acl}

\usepackage{times}
\usepackage{latexsym}
\usepackage{amssymb}

\usepackage[T1]{fontenc}

\usepackage[utf8]{inputenc}

\usepackage{microtype}

\usepackage{inconsolata}

\usepackage{graphicx}
\usepackage{amsmath} 
\usepackage{booktabs}
\usepackage{multirow}
\usepackage[table]{xcolor}
\newcommand{\best}[1]{\textbf{#1}}
\newcommand{\second}[1]{\underline{#1}}
\usepackage[most]{tcolorbox}
\tcbuselibrary{listings,breakable}
\newtcblisting{PromptBlock}{
  breakable,
  enhanced,
  colback=gray!6,
  colframe=gray!60,
  boxrule=0.4pt,
  arc=2pt,
  left=4pt,
  right=4pt,
  top=4pt,
  bottom=4pt,
  listing only,
  listing engine=listings,
  listing options={
    basicstyle=\ttfamily\footnotesize,
    breaklines=true,
    breakatwhitespace=false,
    columns=fullflexible,
    keepspaces=true,
    showstringspaces=false
  }
}

\title{Mitigating Provenance-Role Collapse in Long-Term Agents via \\Typed Memory Representation
}

\author{
  \textbf{Zhengda Jin\textsuperscript{1}}\thanks{Equal contribution.},
  \textbf{Bingbing Wang\textsuperscript{1,2}}\footnotemark[1],
  \textbf{Jing Li\textsuperscript{2}},
  \textbf{Ruifeng Xu\textsuperscript{1,3}}\thanks{Corresponding authors.},
  \textbf{Min Zhang\textsuperscript{1}}
\\
\\
  \textsuperscript{1}Harbin Institute of Technology, Shenzhen, China \\
  \textsuperscript{2}The Hong Kong Polytechnic University, Hong Kong, China \\
  \textsuperscript{3}Shenzhen Loop Area Institute, Shenzhen, China
}

\begin{document}
\maketitle
\begin{abstract}
Long-term memory is essential for persistent LLM agents, yet prevailing architectures store historical interactions as unstructured, flat text. This unconstrained storage induces \textbf{provenance-role collapse}, a critical failure mode where agents suffer from source-monitoring errors. To resolve this cognitive vulnerability at the architectural level, we propose \textbf{\textsc{MemIR}}, a typed \textbf{\textsc{Mem}}ory \textbf{I}ntermediate \textbf{R}epresentation that operationalizes source monitoring as a structural constraint. \textsc{MemIR} writes long-term memory into grounded atoms that separate raw evidence, retrieval cues, and truth-bearing claims, with factual authorization restricted to supported claim atoms. It then applies multi-route atomic projection and provenance-scoped utilization to transform heterogeneous retrieval hits into claim-centered candidate bundles and a normalized fact interface for answer generation. Experiments on LoCoMo and BEAM-100K demonstrate that \textsc{MemIR} consistently outperforms existing memory baselines, especially on tasks requiring source tracking, temporal grounding, and aggregation of fragmented evidence.
\end{abstract}

\section{Introduction}
Long-term memory is transforming Large Language Model (LLM) agents from stateless responders into persistent, personalized assistants \citep{packer2023memgpt,chen2026telemem}. To navigate extended temporal horizons, an agent must accumulate experiential streams and recover relevant information across fragmented sessions \citep{zhong2024memorybank,zhou2023recurrentgpt,sun2026hmem}. This requires a mechanism that can seamlessly recontextualize historical traces into present tasks \citep{lewis2020retrieval,yu2026agentic} while preserving their original narrative boundaries and epistemic roles.

\begin{figure}[!t]
  \centering
  \includegraphics[width=1.0\linewidth]{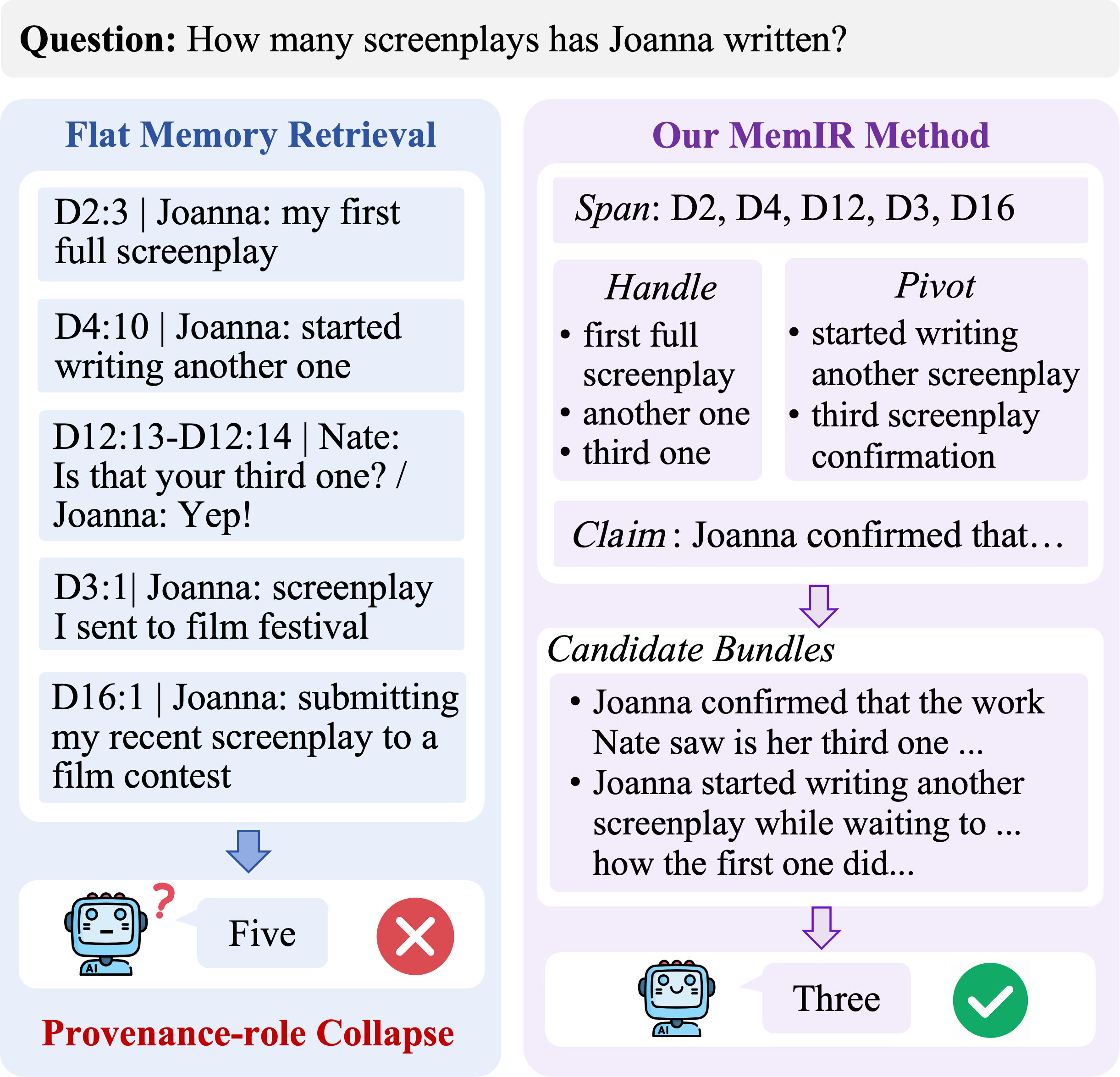}
  \caption{Example of the existing method and our \textsc{MemIR} approach.}
  \label{fig:intro}
\end{figure}

Existing long-term memory architectures predominantly treat memory as an amorphous pool of retrievable flat text, where historical interactions are compressed into untyped summaries or narrative chunks and recalled via lexical or dense retrieval. Such flattening removes the structural cues required for \textit{source monitoring}: distinguishing observed evidence from inferred content, tracking referents across mentions, and identifying the epistemic status of retrieved statements. This leads to \textbf{provenance-role collapse}, a failure mode in which fragments with different provenance or discourse roles are either merged without authorization or treated as independent entities despite referring to the same evolving object. As shown in Figure~\ref{fig:intro}, answering \textit{``How many screenplays has Joanna written?''} requires aggregating fragmented evidence about her first screenplay, a subsequent project, and a recent work; yet flat memory retrieval lack explicit coreference links, source boundaries, and epistemic typing, and may overcount temporally distributed mentions as distinct objects rather than stages of a single referent. Therefore, reliable memory utilization requires not only retrieval efficacy, but a structural mechanism for source monitoring that preserves provenance and epistemic roles.

To bridge this gap, we propose \textbf{\textsc{MemIR}}, a typed \textbf{MEM}ory \textbf{I}ntermediate \textbf{R}epresentation that turns source monitoring into an explicit structural constraint for long-term memory. Instead of storing interaction histories as homogeneous text, \textsc{MemIR} compiles them into grounded \textit{memory atoms} that separate raw evidence, cues, and truth-bearing claims, ensuring that only supported claim atoms can serve as factual memory.
\textsc{MemIR} further organizes retrieval around claim-centered evidence use. Heterogeneous retrieval hits from sparse and dense routes are projected through cross-atom retrieval views into provenance-scoped candidate bundles, where evidence fragments and auxiliary cues can contribute only through their associated claims. The selected bundles are then exposed as a normalized fact interface for answer generation, allowing downstream models to reason over compact, source-grounded factual atoms.
Our main contributions are summarized as follows:

\begin{itemize}
    \item We introduce \textsc{MemIR}, a typed memory intermediate representation that organizes long-term memory as \textbf{memory atoms} to operationalize source monitoring.
    \item We develop multi-route atomic projection and provenance-scoped utilization modules for \textsc{MemIR}, using type-constrained projection to transform heterogeneous retrieval hits into claim-centered candidate bundles and a normalized fact interface for answer generation.
    \item Extensive experiments on LoCoMo and BEAM-100K show that \textsc{MemIR} consistently improves long-term memory reasoning compared with existing memory baselines.
\end{itemize}

\section{Related Work}

\subsection{Long-Horizon Memory Systems}

Long-horizon memory is foundational for enabling persistent behavior in LLM agents~\citep{park2023generative,packer2023memgpt,chen2026telemem}. Recent work has moved from simple retrieval-augmented pipelines to autonomous memory management systems, including reinforcement learning (RL)-driven CRUD operations~\citep{yu2026agentic,yan2025memory}, neuro-symbolic memory palaces~\citep{arslan2026aeon}, and active context compression~\citep{verma2026active,liu2026simplemem}. Evaluation has similarly evolved toward complex multi-turn benchmarks such as LoCoMo~\citep{maharana2024evaluating} and MemoryAgentBench~\citep{hu2025evaluating}. However, most systems still store and retrieve memory as untyped textual artifacts, such as summaries, reflections, or compressed context tokens~\citep{wang2023voyager,zhou2023recurrentgpt}. These flat representations force the model to infer the provenance and epistemic role of retrieved items at generation time, making memory use vulnerable to provenance-role collapse. \textsc{MemIR} addresses this limitation by representing memory as typed, source-grounded atoms and routing retrieved evidence through claim-centered bundles and a provenance-scoped fact interface.

\subsection{Cognitive-Inspired Memory}
To bridge the gap between flat-text retrieval and complex reasoning, prior work has explored structured memory organization. These approaches range from citation-aware retrieval \citep{nakano2021webgpt} to graph-based text indexing \citep{edge2024local,gutierrez2024hipporag,sun2026hmem}. Recent architectures have further incorporated cognitive-inspired mechanisms, such as test-time memorization \citep{behrouz2026titans} and proactive memory extraction \citep{promem2026proactive}. While these methods enhance multi-hop reasoning and access efficiency, structure alone does not specify the functional authority of memory artifacts. 
Citations or referential anchor may help locate relevant context, but it can not be treated as a truth-bearing assertion.

Grounded in cognitive source monitoring \citep{johnson1993source}, \textsc{MemIR} operationalizes memory authority by separating factual authorization from raw retrieval hits. Unlike prior systems that primarily optimize for retrieval relevance or storage compression \citep{hu2023chatdb,edge2024local}, \textsc{MemIR} preserves original provenance roles across the entire memory pipeline. This provides a formal framework for verifiable memory utilization.

\begin{figure*}[!t]
  \centering
  \includegraphics[width=\linewidth]{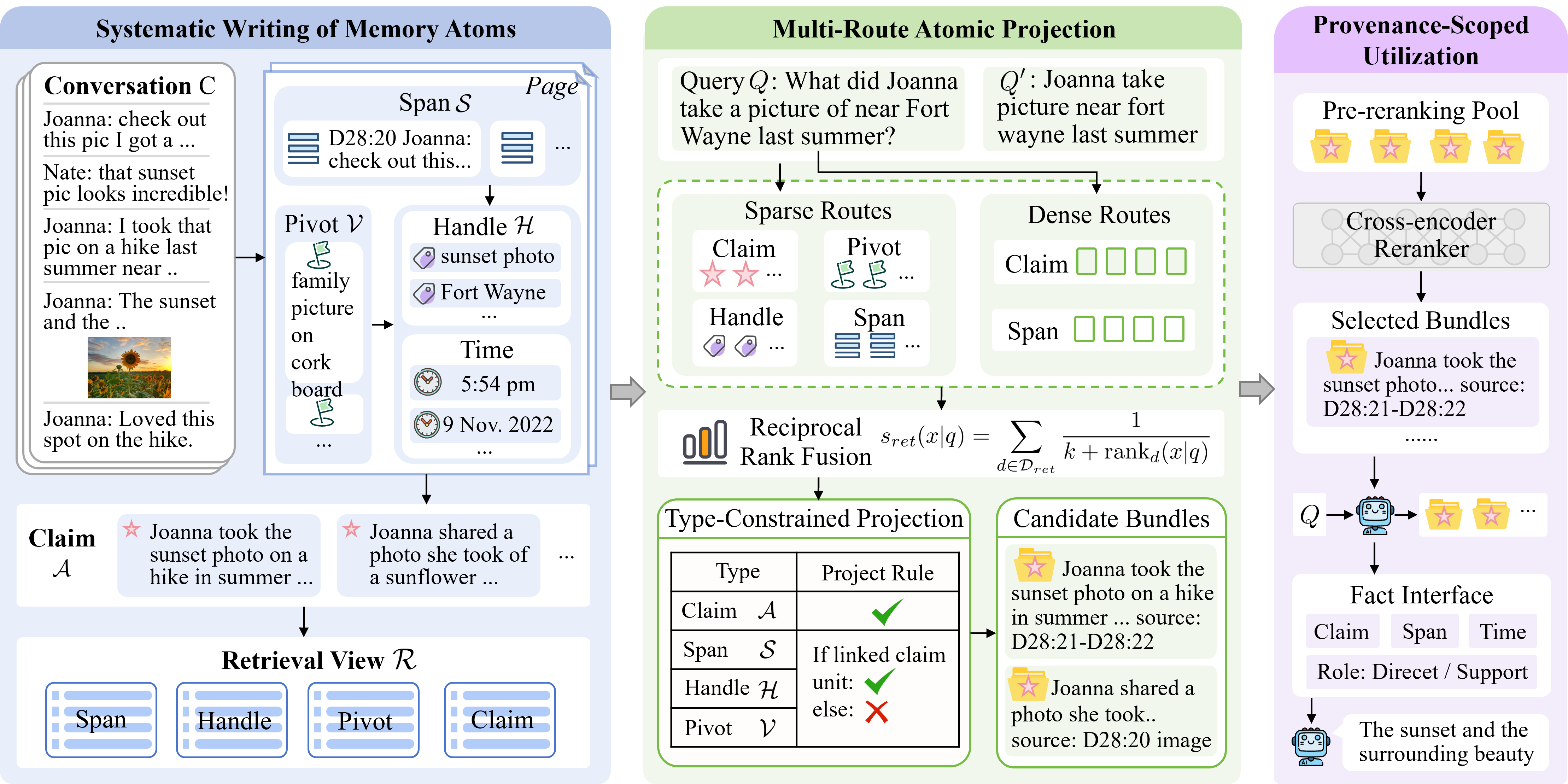}
  \caption{Overview of \textsc{MemIR}, comprising systematic writing of memory atoms, multi-route atomic projection, and provenance-scoped utilization.}
  \label{fig:framework}
\end{figure*}

\section{\textsc{MemIR} Method}

This section presents \textbf{\textsc{Mem}}ory \textbf{I}ntermediate \textbf{R}epresentation (\textbf{\textsc{MemIR}}), a typed schema for managing long-term memory. 
Given an interaction history $\mathcal{C}$, \textsc{MemIR} compiles it into a memory intermediate representation $\mathcal{M}$ that composed role-annotated memory atoms. For a query $q$, \textsc{MemIR} retrieves relevant artifacts from $\mathcal{M}$ and lowers them into a provenance-preserving fact interface $F_q$, which serves as structured evidence for answer generation, yielding $y$. 
As shown in Figure \ref{fig:framework}, \textsc{MemIR} consists of three coupled stages.  First, \textbf{Systematic Writing of Memory Atoms} transforms unstructured interaction history into typed memory atom. Then, \textbf{Multi-Route Atomic Projection} retrieves relevant memory atoms through complementary access routes and projects heterogeneous hits into claim-centered candidates. Ultimately, \textbf{Provenance-Scoped Utilization} reranks and selects the projected candidates, producing a compact fact interface for downstream answer generation.

\subsection{Systematic Writing of Memory Atoms}

To prevent raw evidence, auxiliary cues, and derived facts from being collapsed into homogeneous records, \textsc{MemIR} formulates memory writing as a typed compilation process $\Phi: \mathcal{C} \to \mathcal{M}$. A \textbf{memory atom} is the minimal structured unit that can be independently stored, indexed, and retrieved. The resulting memory store is defined as $\mathcal{M} = \{ \mathcal{P}, \mathcal{S}, \mathcal{H}, \mathcal{T}, \mathcal{V}, \mathcal{A}, \mathcal{R}\}$, where $\mathcal{P}$ and $\mathcal{S}$ denote \textit{page} and \textit{span} atoms for preserving local context boundaries and verbatim evidence; $\mathcal{H}$, $\mathcal{T}$, and $\mathcal{V}$ denote \textit{handle}, \textit{time}, and \textit{pivot} atoms for referential, temporal, and semantic access. $\mathcal{A}$ denotes \textit{claim} atoms for source-grounded factual assertions, and $\mathcal{R}$ denotes \textit{retrieval views} that expose atoms to different search mechanisms. 

The compilation $\Phi$ first discretizes the raw context into a grounding substrate by grouping dialogue turns into page $\mathcal{P}$ and segmenting them into span $\mathcal{S}$ containing verbatim evidence. Based on these atoms, an LLM-powered cue extraction function $f_{\mathrm{cue}}: \mathcal{S} \to \mathcal{H} \cup \mathcal{T} \cup \mathcal{V}$ identifies auxiliary \textit{handle}, \textit{time}, and \textit{pivot} atoms for referential and temporal anchoring. 
Motivated by source monitoring theory \cite{johnson1993source}, which emphasizes distinguishing externally observed evidence from internally generated or reconstructed content, \textsc{MemIR} operationalizes this distinction through a strict support constraint:
$\varnothing \neq \operatorname{sup}(x) \subseteq \mathcal{S}, \quad x \notin \mathcal{P}$.
Here, $\operatorname{sup}(x)$ denotes the \textit{span} atoms that ground $x$. This constraint requires every extracted cue or generated claim to retain an explicit evidential origin, preventing LLM-derived memory atoms from being mistaken for unsupported facts.

For each \textit{page} atom $P \in \mathcal{P}$, a compilation function
$f_{\mathrm{write}}(P, \mathcal{H}, \mathcal{T}, \mathcal{V}) \to \mathcal{A}_P$
generates source-grounded \textit{claim} atoms via an LLM, which represent the only truth-bearing components within $\mathcal{M}$. Finally, to decouple physical storage from logical access, a mapping $f_{\mathrm{view}}$ constructs a set of \textit{retrieval views} $\mathcal{R}_x$ for each atom $x$. Rather than creating new memory content, a \textit{retrieval view} specifies cross-atom access relations, such as linking a span to the \textit{claims} it supports, a \textit{handle} to the \textit{claims} and \textit{spans} in which the referent appears, or a \textit{pivot/time} atom to the \textit{claims} sharing the same semantic or temporal anchor. These views allow retrieval to traverse heterogeneous atom types while preserving claim-level factual authorization.

\subsection{Multi-Route Atomic Projection}
To support diverse memory access paths, \textsc{MemIR} implements a multi-route projection mechanism that retrieves memory atoms through heterogeneous access routes. Rather than treating retrieved fragments as an independent context, \textsc{MemIR} first identifies potentially relevant atoms and then projects these heterogeneous hits into a claim-centered candidate space.

Given a query $q$, \textsc{MemIR} identifies candidate memory atoms through retrieval routes $\mathcal{D}_{\mathrm{ret}}$ that combine sparse lexical matching with dense semantic retrieval. Sparse routes use BM25 over retrieval views $\mathcal{R}$ to access atoms through lexical keys, referential aliases, temporal expressions, and local-context expansions, covering the views of \textit{claim}, \textit{span}, \textit{handle}, \textit{time}, and \textit{pivot} atoms. Before sparse retrieval, the query is rewritten into surface forms $\mathcal{Q}(q)$ via a function-word table to better match view-based keys; view-level hits are then merged into atom-level hits for structural consistency. Dense routes use BGE-M3~\citep{chen2024m3} over \textit{claim} and \textit{span}, where atoms contain sufficient natural-language content for semantic matching. These routes complement sparse retrieval by capturing paraphrases and weak lexical overlaps that may not match symbolic views.

Let $I_d$ be the index associated with route $d \in \mathcal{D}_{\mathrm{ret}}$. The initial hit set is defined as the union of top-$K$ results across all routes: $\mathcal{X}_q = \bigcup_{d \in \mathcal{D}_{\mathrm{ret}}} \operatorname{TopK}_d(q, I_d)$. To combine ranking signals, we apply Reciprocal Rank Fusion (RRF) \citep{cormack2009reciprocal}:
\begin{equation}
s_{\mathrm{ret}}(x \mid q) = \sum_{d \in \mathcal{D}_{\mathrm{ret}}} \frac{\mathbf{1}[x \in \operatorname{TopK}_d(q, I_d)]}{k + \operatorname{rank}_d(x \mid q)}
\end{equation}
where $k$ is a smoothing constant and $\operatorname{rank}_d$ denotes the rank of memory atom $x$ within route $d$.

To bridge the gap between auxiliary evidence and factual candidates, \textsc{MemIR} applies a type-constrained projection function that maps heterogeneous hits into the set of \textit{claim} atoms $\mathcal{A}$. Let $o(h)$ denote the underlying memory atom of hit $h \in \mathcal{X}_q$. The projection rule is defined as:
\begin{equation}
\Pi(h) = \{ a \in \mathcal{A} \mid o(h) \in \Omega_a \cup \{a\} \}
\end{equation}
where $\Omega_a$ represents the association set of claim $a$, encompassing its supporting spans $\operatorname{sup}(a)$ and associated referential cues (\textit{handles} and \textit{pivots}). This unified mapping ensures that \textit{span}, \textit{handle}, and \textit{pivot} atoms enter the factual candidate layer exclusively through their linked claims, while hits without a valid claim association are discarded. 

Finally, all hits projecting to the same claim are consolidated into a {candidate bundle} $b_a = \langle a, \rho_a, \mathcal{E}_a \rangle$. Specifically, let $\mathcal{H}_a = \{ h \in \mathcal{X}_q \mid a \in \Pi(h) \}$ be the set of hits supporting claim $a$. The aggregated retrieval strength $\rho_a$ is computed as the sum of fusion scores across its supporting hits:
\begin{equation}
\rho_a = \sum_{h \in \mathcal{H}_a} s_{\mathrm{ret}}(o(h) \mid q)
\end{equation}
where $\mathcal{E}_a = \{ o(h) \mid h \in \mathcal{H}_a \} \cup \Omega_a$ represents the \textit{provenance closure}. This closure encapsulates both retrieved evidence and the claim's complete association set, providing a self-contained factual unit for subsequent reranking and grounded generation.

\begin{table*}[!t]
\centering
\scriptsize
\setlength{\tabcolsep}{3.0pt}
\renewcommand{\arraystretch}{1.05}
\resizebox{\textwidth}{!}{
\begin{tabular}{lcccccccccccc}
\toprule
\multirow{2}{*}{\textbf{Method}} 
& \multicolumn{3}{c}{\textbf{Single-hop}}
& \multicolumn{3}{c}{\textbf{Multi-hop}}
& \multicolumn{3}{c}{\textbf{Temporal}}
& \multicolumn{3}{c}{\textbf{Open-domain}} \\
\cmidrule(lr){2-4} \cmidrule(lr){5-7} \cmidrule(lr){8-10} \cmidrule(lr){11-13}
& F1 & BLEU-1 & Judge & F1 & BLEU-1 & Judge & F1 & BLEU-1 & Judge & F1 & BLEU-1 & Judge \\
\midrule
\rowcolor{violet!10}
\multicolumn{13}{c}{\textit{\textbf{Backbone: GPT-4.1-mini}}} \\
ReadAgent & 9.20 & 7.90 & - & 6.50 & 5.60 & - & 5.30 & 4.20 & - & 7.70 & 6.60 & - \\ 
LoCoMo & 18.70 & 15.90 & - & 25.00 & 21.60 & - & 12.00 & 10.60 & - & 19.10 & 17.10 & - \\
Zep & 45.50 & 40.00 & 66.90 & 30.50 & 20.40 & 53.70 & 23.90 & 20.00 & 60.20 & 24.20 & 19.30 & 43.80 \\
A-Mem & 45.00 & 39.80 & 64.00 & 30.40 & 20.00 & 55.70 & 40.30 & 33.70 & 66.70 & 13.40 & 12.70 & 37.50 \\
LightMem & 33.80 & 29.70 & 72.50 & 24.90 & 21.70 & 69.60 & 20.60 & 18.40 & 68.10 & 19.20 & 17.70 & 52.40 \\
MemoryOS & 44.20 & 37.50 & 68.90 & 34.00 & 25.80 & 62.40 & 36.50 & 27.40 & 37.70 & 30.20 & 25.60 & \second{60.40} \\
Mem0 & 48.60 & 42.00 & 71.40 & 40.10 & 30.30 & 68.20 & 40.30 & 33.70 & 56.90 & 23.70 & 17.70 & 47.90 \\
SwiftMem & 45.50 & 48.00 & 76.70 & 32.00& \best{37.10} & 58.90  & 50.70 & \second{56.90} & 68.50 & 25.80 & \best{29.90} & 55.20 \\
HIMEM & 49.10 & 43.50 & 74.30 & 30.30 & 23.70 & 71.50 & 43.00 & 38.80 & 79.20 & 21.90 & 10.20 & 43.40 \\
LangMem & 51.00 & 43.60 & 84.50 & \second{41.50} & 32.50 & 71.00 & 48.50 & 40.90 & 50.80 & \second{32.80} & 26.40 & 59.00 \\
NEMORI & 55.70 & 49.50 & 87.00 & 40.80 & 31.70 & \best{74.80} & 58.70 & 50.70 & 77.30 & 31.70 & 25.10 & 56.30 \\
SimpleMem & \second{57.70} & \second{51.60} & \second{87.40} & 39.60 & 31.50 & \second{73.80} & \second{60.40} & 53.40 & \second{80.70} & 26.90 & 21.30 & 57.30 \\
\rowcolor{gray!10}
\textbf{\textsc{MemIR} (Ours)} & \best{59.60} & \best{54.20} & \best{89.50} & \best{42.50} & \second{34.10} & 70.20 & \best{63.50} & \best{58.70} & \best{84.60} & \best{33.10} & \second{28.10} & \best{61.90} \\
\midrule
\rowcolor{violet!10}
\multicolumn{13}{c}{\textit{\textbf{Backbone: GPT-4.1}}} \\
LightMem & 32.00 & 29.50 & 71.40 & 28.70 & 25.60 & 68.30 & 24.60 & 20.40 & 66.20 & 23.80 & 19.90 & 50.20 \\
MemoryOS & 43.10 & 36.80 & 65.90 & 35.90 & 26.00 & 66.40 & 39.80 & 28.50 & 42.70 & 30.70 & 26.00 & 54.60 \\
HIMEM & 45.60 & 40.70 & 70.80 & 33.80 & 25.70 & 69.40 & 46.80 & 39.60 & 80.50 & 20.60 & 9.00  & 38.20 \\
A-Mem & 46.50 & 40.10 & 62.10 & 26.70 & 18.40 & 56.30 & 42.20 & 33.60 & 68.90 & 14.20 & 15.40 & 40.50 \\
Mem0 & 50.40 & 43.20 & 75.90 & 41.30 & 32.50 & 70.30 & 43.50 & 36.70 & 60.70 & 23.40 & 17.40 & 45.10 \\
LangMem &51.13 &44.22&-& 41.11 &32.09&-& 53.67& 46.22&-& \second{33.38} &\second{27.26}&-\\
NEMORI & 53.80 & 45.40 & \second{85.60} & 40.10 & 32.60 & \second{74.00} & 59.20 & 52.40 & 79.60 & 30.20 & 20.50 & 51.90 \\
O-Mem&54.89 &48.98&-&\second{42.64}& 34.08&-&57.48 &49.76 &-&30.58 &25.69&-\\
SimpleMem & \second{55.80} & \second{49.90} & 83.80 & 42.30 & \second{34.40} & \best{75.20} & \second{60.10} & \second{53.60} & \second{81.90} & 28.30 & 22.60 & \second{55.20} \\

\rowcolor{gray!10}
\textbf{\textsc{MemIR} (Ours)} & \best{62.20} & \best{56.70} & \best{86.80} & \best{42.90} & \best{35.00} & 71.70 & \best{61.80} & \best{55.70} & \best{86.30} & \best{34.20} & \best{28.20} & \best{58.70} \\

\bottomrule
\end{tabular}
}
\caption{Main results on LoCoMo. The \textbf{bold} and \underline{underlined} values denote the best and second-best results within each backbone group, respectively.} 
\vspace{-3.5pt}
\label{tab:locomo-main}
\end{table*}

\subsection{Provenance-Scoped Utilization}
To resolve redundancy and identify task-relevant facts, we construct a compact, provenance-scoped input through a coarse-to-fine reranking and selection process. \textsc{MemIR} first ranks candidate bundles by their aggregated retrieval scores $\rho_a$ and retains the top-$M$ bundles as a pre-reranking pool $\mathcal{B}^{(M)}_q$, maintaining broad coverage of potentially relevant claims. To model finer-grained query-fact interactions, a cross-encoder $g_\theta$ then scores each bundle $b_a \in \mathcal{B}^{(M)}_q$:
$s_{\mathrm{rank}}(b_a \mid q) = g_\theta(q, \psi_{\mathrm{rank}}(b_a))$,
where $\psi_{\mathrm{rank}}(b_a)$ denotes a textual serialization of the bundle. From the top-$K$ reranked bundles $(K \leq M)$, an LLM selector further chooses at most $X$ complementary candidates
$\widehat{\mathcal{B}}_q = \{ (b_i, r_i) \}_{i=1}^m$, where $m \leq X$ and $r_i \in \{\text{direct}, \text{support}\}$ denotes the functional role of bundle $b_i$ in answering $q$.

The selected bundles $\widehat{\mathcal{B}}_q$ are finally transformed into a normalized fact interface $F_q = \{ f_1, \dots, f_m \}$, where each selected item is $(b_i, r_i) \in \widehat{\mathcal{B}}_q$ with $b_i = \langle a_i, \rho_i, \mathcal{E}_i \rangle$, and each fact record is defined as $f_i = \langle z(a_i), \mathcal{E}_i, \tau_i, r_i \rangle.$ Here, $z(a_i)$ denotes the factual text of claim atom $a_i$, $\mathcal{E}_i$ is its provenance closure, $\tau_i$ denotes the temporal cues associated with $a_i$ as exposed by $\mathcal{E}_i$, and $r_i$ is the assigned role. This structured interface acts as the primary input to the answer model $f_\phi: y = f_\phi(q, F_q)$, where $y$ is the generated response. By receiving source-grounded fact records rather than raw retrieval hits, the answer model is forced to operate within a provenance-scoped boundary. Crucially, if $F_q$ contains insufficient evidence to fulfill the query requirements, $f_\phi$ is instructed to return an "\textit{insufficient evidence}" instead of engaging in unsupported generation, so as to mitigate hallucinations.

\section{Experiment}
\subsection{Experimental Setup}
\paragraph{Datasets and Metrics.} 
We evaluate \textsc{MemIR} on two representative long-term memory benchmarks. 1) \textbf{LoCoMo}~\cite{maharana2024evaluating} contains 10 long multi-turn dialogues and 1,540 questions covering single-hop, multi-hop, temporal, and open-domain question answering. Following Mem0~\cite{chhikara2025mem0}, we report token-level F1 (\textbf{F1}), \textbf{BLEU-1}, and LLM-based evaluation scores (\textbf{Judge}) for the four answerable categories using the same evaluation prompt.
2) \textbf{BEAM-100K} \cite{tavakoli2026beyond}, which contains 20 dialogues with 100K-token histories and 400 questions across ten task categories. Following the original protocol, we report \textbf{LLM-as-Judge} scores for each category and use the official category abbreviations in the results tables. 
Details are shown in Appendix~\ref{app:data_metric}.

\paragraph{Implementation Details.}
For LoCoMo, we report results using GPT-4.1-mini and GPT-4.1 as backbone models. Unless otherwise specified, all BEAM-100K experiments use GPT-4.1-mini for memory construction, bundle selection, and answer generation, while LLM-as-Judge evaluation is performed with GPT-4o-mini. Full configuration details are listed in Appendix~\ref{app:implementation_detail}.

\subsection{Baselines}
We compare \textsc{MemIR} with representative long-term memory systems: LoCoMo \cite{maharana2024evaluating}, ReadAgent \cite{lee2024human}, Zep\cite{rasmussen2025zep}, LangMem\footnote{https://github.com/langchain-ai/langmem.}, A-Mem \cite{xu2026mem}, MemoryOS \cite{kang2025memory}, Mem0 \cite{chhikara2025mem0}, LightMem \cite{fang2026lightmem}; NEMORI \cite{ma2025deserves}, SimpleMem \cite{liu2026simplemem}, HIMEM \cite{zhang2026himem}, O-Mem \cite{wang2025mem}, and SwiftMem \cite{tian2026swiftmem}. Detailed descriptions of each baseline are provided in Appendix~\ref{app:baselines}.

\begin{table*}[!t]
\centering
\scriptsize
\setlength{\tabcolsep}{3.8pt}
\renewcommand{\arraystretch}{1.05}
\resizebox{\textwidth}{!}{
\begin{tabular}{lccccccccccc}
\toprule
\textbf{Method}
& \textbf{ABS} & \textbf{CR} & \textbf{EO} & \textbf{IE} & \textbf{IF}
& \textbf{KU} & \textbf{MSR} & \textbf{PF} & \textbf{SUM} & \textbf{TR}
& \textbf{Avg.} \\
\midrule
MemoryOS
& 5.00  & 10.50 & 12.70 & 36.40 & 52.60
& 40.90 & 25.20 & 54.50 & 22.40 & 26.20
& 28.64 \\

LightMem
& 10.00  & 15.40 & 18.10 & 50.80 & 45.60
& 41.60 & 30.00 & 59.80 & 28.50 & 24.80
& 32.46 \\

Mem0
& 20.00  & 18.40 & 22.60 & 58.40 & 57.80
& 45.10 & 33.30 & 52.50 & 30.90 & 31.50
& 37.05 \\

NEMORI
& \second{25.00}  & 18.90 & 20.40 & \second{63.50} & \second{64.50}
& 52.10 & 32.40 & 64.00 & \second{31.70} & 32.00
& 40.45 \\

SimpleMem
& 12.50 & \second{20.60} & \second{25.10} & 60.70 & 60.00
& \second{55.00} & \second{38.10} & \second{68.10} & 30.70 & \second{35.00}
& \second{40.68} \\

\rowcolor{gray!10}
\textbf{\textsc{MemIR} (Ours)}
& \best{37.50} & \best{32.30} & \best{28.60} & \best{69.80} & \best{65.60}
& \best{58.40} & \best{43.70} & \best{71.40} & \best{36.80} & \best{38.50}
& \best{48.26} \\
\bottomrule
\end{tabular}
}
\caption{Main results on BEAM-100K with GPT-4.1-mini.  We report LLM-as-Judge scores for ABS, CR, EO, IE, IF, KU, MSR, PF, SUM, and TR, corresponding to abstention, contradiction resolution, event ordering, information extraction, instruction following, knowledge update, multi-session reasoning, preference following, summarization, and temporal reasoning. Avg. is the category average. Best and second-best scores are \textbf{bolded} and \underline{underlined}.}
\label{tab:beam-main}
\end{table*}

\begin{figure*}[!t]
  \centering
  \includegraphics[width=1.0\linewidth]{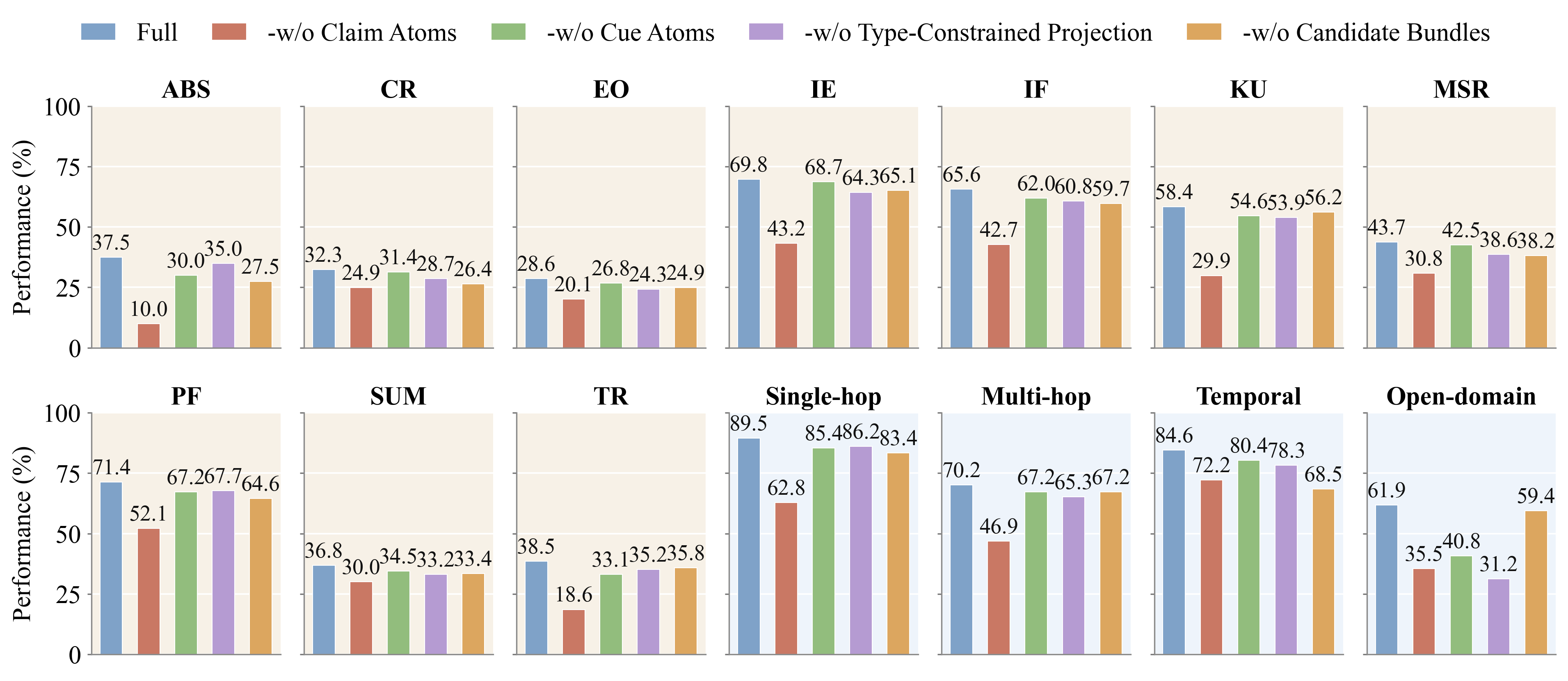}
  \caption{Ablation results on BEAM-100K and LoCoMo with GPT-4.1-mini. Each subplot reports the performance of the full \textsc{MemIR} model and four ablated variants. BEAM-100K subplots show LLM-as-Judge scores for ten task categories, while LoCoMo subplots show category-level average scores over F1, BLEU-1, and LLM-as-Judge for Single-hop, Multi-hop, Temporal, and Open-domain questions.}
  \label{fig:ablation}
\end{figure*}

\subsection{Main Results}
The results on LoCoMo and BEAM-100K are illustrated in Table~\ref{tab:locomo-main} and \ref{tab:beam-main}. On LoCoMo, \textsc{MemIR} achieves the strongest overall performance under both GPT-4.1-mini and GPT-4.1, with the clearest gains on single-hop, temporal, and open-domain questions. This pattern is consistent with the design of \textsc{MemIR}: when the key challenge is to recover a specific supported fact, preserve its temporal anchoring, or expose it in a directly consumable form, the claim-centered representation provides a clear advantage over flat memory text. By contrast, The smaller advantage on multi-hop LLM-judge scores indicates that \textsc{MemIR} improves evidence organization, while complex cross-claim reasoning remains a challenge for the downstream reasoner.

BEAM-100K further shows this advantage persists under substantially longer histories and heterogeneous memory-use demands. \textsc{MemIR} outperforms all baselines, with strong gains on contradiction resolution, knowledge update, multi-session/temporal reasoning, and summarization. These tasks emphasize tracking evolving states and integrating distant evidence, precisely where provenance boundaries and claim-level factual authorization matter most. Taken together, the results suggest the benefit of \textsc{MemIR} does not arise merely from stronger retrieval coverage, but from converting long-horizon interaction history into provenance-scoped factual units that remain easier to access, compare, and consume at answer time.

\begin{table*}[t]
\centering
\setlength{\tabcolsep}{2.0pt} 
\renewcommand{\arraystretch}{1.05} 
\resizebox{\textwidth}{!}{
\begin{tabular}{lcccccccccccccc}
\toprule
\multirow{2}{*}{\textbf{Model}} 
& \multicolumn{10}{c}{\textbf{BEAM-100K}} 
& \multicolumn{4}{c}{\textbf{LoCoMo}} \\
\cmidrule(lr){2-11} \cmidrule(lr){12-15}
& \textbf{ABS} 
& \textbf{CR} 
& \textbf{EO} 
& \textbf{IE} 
& \textbf{IF} 
& \textbf{KU} 
& \textbf{MSR} 
& \textbf{PF} 
& \textbf{SUM} 
& \textbf{TR} 
& \textbf{Single-Hop} 
& \textbf{Multi-hop} 
& \textbf{Temporal} 
& \textbf{Open-domain} \\
\midrule
GPT-4.1 
& \best{40.0} & \best{36.7} & \best{29.8} & 64.7 & \best{66.2} & \best{60.5} & \best{48.1} & \best{72.7} & \best{40.6} & 36.8 
& 86.8 & \best{71.7} & \best{86.3} & 58.7 \\
GPT-4.1-mini 
& 37.5 & 32.3 & 28.6 & \best{69.8} & 65.6 & 58.4 & 43.7 & 71.4 & 36.8 & \best{38.5} 
& \best{89.5} & 70.2 & 84.6 & \best{61.9} \\
Qwen3-14B 
& 15.0 & 21.2 & {19.7} & 47.9 & 40.8 & 38.4 & 20.7 & 54.9 & 19.8 & 21.0 
& 79.1 & 62.0 & 58.6 & 31.2 \\
MiMo-7B 
& 17.5 & 12.4 & 9.7 & 26.9 & 24.8 & 21.2 & 11.5 & 34.7 & 8.7 & 12.2 
& 46.0 & 52.8 & 32.1 & 37.5 \\
GLM-4 9B 
& 10.0 & 18.6 & 14.6 & 28.5 & 31.7 & 33.4 & 14.3 & 35.8 & 17.9 & 20.4 
& 68.9 & 47.9 & 57.7 & 19.6 \\
\bottomrule
\end{tabular}
}
\caption{Performance of different backbone models on BEAM-100K and LoCoMo.}
\label{tab:backbone}
\end{table*}

\begin{figure*}[!t]
  \centering
  \includegraphics[width=1.0\linewidth]{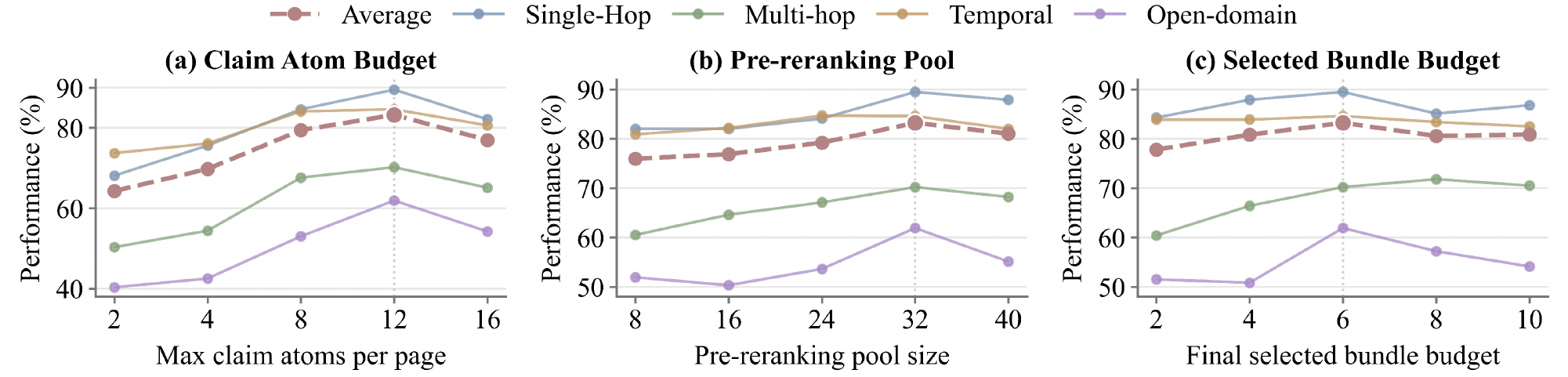}
  \caption{Hyperparameter analysis of \textsc{MemIR} on LoCoMo under different claim atom budgets, pre-reranking pool sizes $M$, and selected bundle budgets $X$.
The vertical dotted line denotes the default setting.}
  \label{fig:hyperparameter-analysis}
\end{figure*}

\subsection{Ablation Study}

We conduct ablation experiments on LoCoMo and BEAM-100K to isolate the contribution of the main structural components in \textsc{MemIR}. As shown in Figure~\ref{fig:ablation}, \textit{w/o Claim Atoms} removes the claim-writing stage and makes the model answer from retrieved span atoms or page-level context, without an explicit truth-bearing factual layer. \textit{w/o Cue Atoms} removes handle, time, and pivot atoms, while keeping claim atoms and their supporting spans. \textit{w/o Type-Constrained Projection} disables the projection from heterogeneous retrieval hits to associated claim atoms, allowing retrieved atoms to enter the candidate layer without claim-level factual authorization. \textit{w/o Candidate Bundles} removes the claim-centered bundle structure and flattens selected claims, evidence spans, cue atoms, and retrieval paths before answer generation.

Across benchmarks, removing claim atoms forces models to infer facts directly from retrieved fragments, blurring topical relevance with factual validity. Omitting cue atoms weakens referential and temporal grounding, complicating the disambiguation of entities and timelines. Disabling type-constrained projection breaks the evidence-fact authorization boundary, allowing auxiliary hits to bypass supported claims. Finally, removing candidate bundles isolates claims from their provenance closure. Ultimately, \textsc{MemIR} improves reliability by explicitly separating factual writing, grounding, authorization, and provenance-scoped consumption.

\subsection{Analysis of Different Backbone}
As shown in Table~\ref{tab:backbone}, the backbone comparison indicates that \textsc{MemIR} contributes a consistent representational benefit, but does not eliminate the need for strong downstream reasoning. Its typed, source-grounded interface appears sufficient to support relatively direct factual access across model families, which helps keep stronger and mid-sized backbones aligned on simpler retrieval-intensive settings. The larger gaps emerge when the answer requires temporal reconciliation, conflict resolution, or synthesis across multiple evidence units, where the bottleneck shifts from memory access to evidence interpretation and controlled generation. This suggests that \textsc{MemIR} primarily improves how long-term evidence is organized and exposed, while the final ability to resolve competing facts, compose dispersed clues, and calibrate answer boundaries still depends heavily on backbone capability.

\begin{figure*}[!t]
  \centering
  \includegraphics[width=1.0\linewidth]{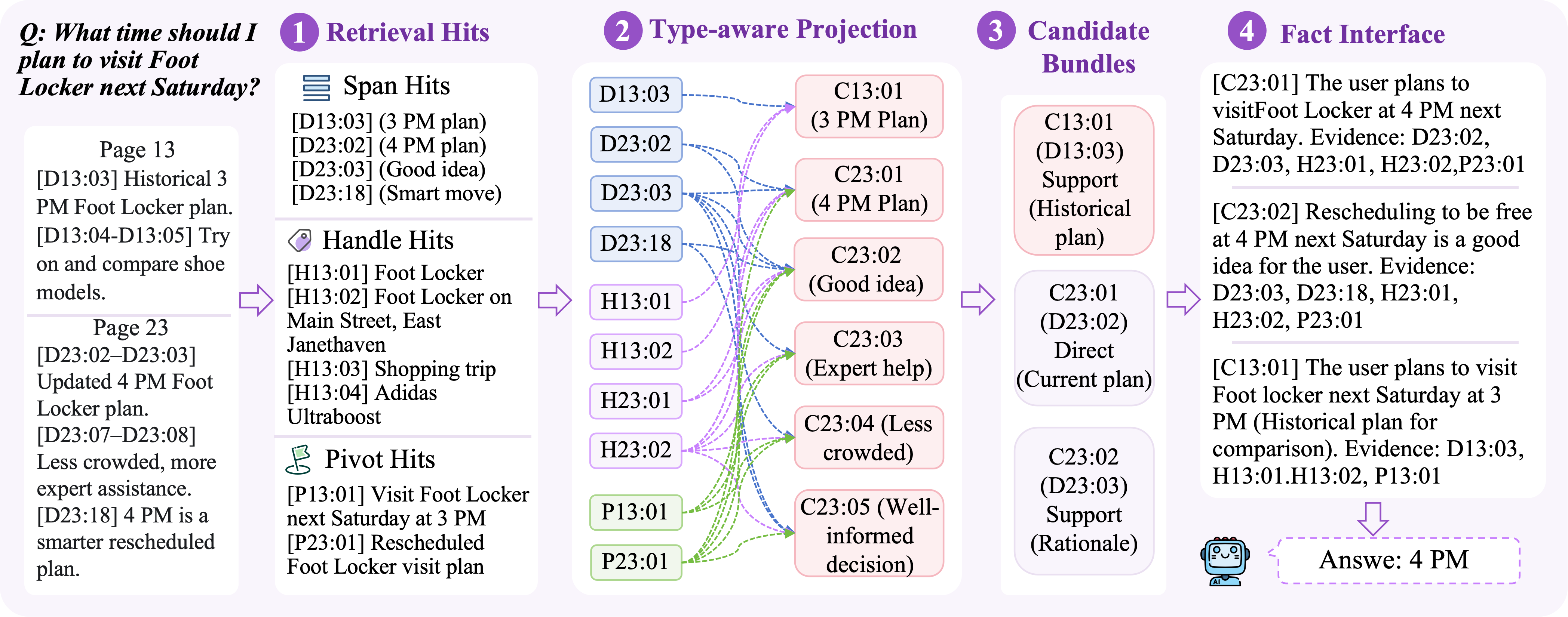}
  \caption{Case Study of our \textsc{MemIR} on BEAM dataset.}
  \label{fig:case}
\end{figure*}

\subsection{Hyperparameter Analysis}
We conduct a hyperparameter analysis on LoCoMo to examine the effect of three structural parameters in \textsc{MemIR}: the maximum number of claim atoms generated per page, the pre-reranking pool size $M$, and the final selected bundle budget $X$. These parameters respectively control the density of source-grounded memory writing, the breadth of claim-centered candidate retrieval before reranking, and the amount of provenance-scoped evidence exposed to the answer model. Figure~\ref{fig:hyperparameter-analysis} reports the overall average performance as well as the performance on different query types.

\paragraph{Claim atom budget.}
When the budget is small, performance is limited because the memory-writing stage cannot sufficiently cover the facts in the source pages. Increasing the budget consistently improves performance across most query types, with the best average result achieved around 12 claim atoms per page. However, further increasing the budget leads to a performance drop, suggesting overly dense claim generation may introduce redundant or overlapping claims, which weakens the precision of the downstream fact interface.

\paragraph{Pre-reranking pool size.}
A larger pool allows the system to collect a broader set of claim-centered candidate bundles before reranking, improving the inclusion of relevant evidence. The performance generally improves as $M$ increases from 8 to 32, especially for the average score and several query categories. Beyond this point, the gains saturate or slightly decline, indicating that once the main relevant bundles are retrieved, adding more candidates mostly increases noise rather than useful coverage.

\paragraph{Selected bundle budget.}
Selecting too few bundles restricts the answer model's access to complementary evidence, particularly harmful for questions requiring multi-evidence aggregation. Increasing $X$ improves performance up to a moderate budget, with the best average performance observed around $X=6$. When more bundles are selected, performance does not continue to improve and may decrease for several query types, suggesting excessive provenance closures can introduce irrelevant facts into the final evidence interface.

\subsection{Case Study}
We further demonstrate a case study in Figure \ref{fig:case}. For the question, retrieval returns two competing memory states: an earlier 3 PM plan from Page 13 and a later 4 PM update from Page 23. This shows that retrieval relevance alone cannot determine the currently valid fact when obsolete and updated evidence are recalled together. \textsc{MemIR} first applies type-aware projection to map heterogeneous hits, including spans, handles, and pivots, onto corresponding Claims, converting raw snippets into fact-bearing candidates. It then performs candidate bundle selection, using C23:01 as direct evidence for the 4 PM answer, C23:02 as supporting rationale, and C13:01 only as historical contrast. Finally, fact interface construction organizes the selected Claims with provenance, enabling the model to answer based on the updated fact. This case highlights the complementary roles of \textsc{MemIR}: projection normalizes retrieval signals, bundle selection resolves conflicting claims, and fact interface construction provides reliable grounded context.

\section{Conclusion}
Long-term memory is essential for persistent LLM agents, but flat-text memory stores obscure provenance, role, and factual authority of retrieved content, causing provenance-role collapse. We introduced \textsc{MemIR}, a typed memory intermediate representation that operationalizes source monitoring through grounded memory atoms, claim-level factual authorization, multi-route atomic projection, and provenance-scoped fact interfaces. Experiments on LoCoMo and BEAM-100K show consistent gains over existing memory baselines. These results suggest reliable long-term memory requires not only effective retrieval, but also explicit provenance preservation and factual authorization.

\section*{Limitations}
\textsc{MemIR} successfully introduces a typed intermediate representation that preserves provenance roles and delivers superior performance on long-horizon memory benchmarks. However, our framework primarily optimizes for evidence structuring and organization, leaving complex cross-claim reasoning heavily dependent on the downstream generation model. Additionally, the systematic writing of multi-type memory atoms inevitably incurs additional computational costs during interaction compilation. Therefore, future work will investigate more lightweight atomic compilation techniques to minimize overhead, and explore explicit structural reasoning mechanisms over the claim-centered bundles to enhance multi-hop logic.




\bibliography{custom}

\appendix

\section{Experimental Setting}
\label{app:appendix_exp_set}
\subsection{Datasets and Metrics}
\label{app:data_metric}
\paragraph{Datasets and Evaluation Metrics.}
We evaluate \textsc{MemIR} on two representative long-term memory benchmarks: LoCoMo~\cite{maharana2024evaluating} and BEAM-100K~\cite{tavakoli2026beyond}.

\textbf{LoCoMo.}
LoCoMo contains 10 long multi-turn dialogues and 1,540 questions spanning four answerable categories: single-hop, multi-hop, temporal, and open-domain question answering. Following Mem0~\cite{chhikara2025mem0}, we report token-level F1, BLEU-1, and LLM-based evaluation scores. The lexical metrics measure surface-level overlap with the reference answers, while the LLM-based metric captures semantic correctness under paraphrased or partially rephrased responses. For LLM-based evaluation, we use \texttt{[GPT-4o-mini} as the judge model and adopt the same evaluation prompt as Mem0. The judge receives the question, the reference answer, and the predicted answer, and determines whether the prediction is semantically consistent with the reference. All compared methods are evaluated with the same judge model and prompt.

\textbf{BEAM-100K.}
BEAM-100K evaluates long-term memory systems under extended contexts and diverse task requirements. It consists of 20 dialogues with 100K-token histories and 400 questions across ten categories: abstention (ABS), contradiction resolution (CR), event ordering (EO), information extraction (IE), instruction following (IF), knowledge update (KU), multi-session reasoning (MSR), preference following (PF), summarization (SUM), and temporal reasoning (TR). Following the original BEAM-100K protocol, we report LLM-as-Judge scores for each category. We use \texttt{[GPT-4o-mini]} as the judge model and keep the original prompt template and scoring criteria unchanged. The judge evaluates each system output according to the category-specific rubric, allowing consistent assessment across heterogeneous reasoning tasks.

\subsection{Implementation Details}
\label{app:implementation_detail}

We report our main results on the LoCoMo benchmark using both GPT-4.1-mini and GPT-4.1 as backbone models. Unless otherwise specified, GPT-4.1-mini is employed for memory construction, bundle selection, and answer generation in all BEAM experiments, ablation studies, and hyperparameter analyses. For the LLM-as-Judge evaluation, we utilize GPT-4o-mini. In our \textsc{MemIR} framework, dense retrieval is performed using bge-m3 \cite{chen2024m3} with FAISS, followed by bundle reranking with bge-reranker-v2-m3 \cite{chen2024m3}. Regarding the hyperparameter configurations, the default setting for LoCoMo generates up to 12 claims per page, retains 32 candidate bundles before both reranking and LLM selection, and ultimately uses the top-6 bundles for final answer generation. To accommodate the significantly longer dialogue contexts in BEAM, we adjust its settings to generate up to 18 claims per page, retain 72 candidate bundles before reranking and selection, and utilize the top-10 bundles to generate the final answer.

\subsection{Baselines}
\label{app:baselines}

We briefly summarize the compared baselines below. 

\textbf{LoCoMo}\cite{maharana2024evaluating}. LoCoMo is primarily a benchmark for evaluating very long-term conversational memory, built from long multi-session dialogues grounded in personas and temporal event graphs. In our comparison, ``LoCoMo'' refers to the reference long-context and retrieval-based baseline setting reported in the LoCoMo paper, rather than the dataset itself.

\textbf{ReadAgent}\cite{lee2024human}. ReadAgent is a human-inspired reading agent for very long contexts. It organizes long inputs into compact gist memories and selectively revisits original passages when finer-grained evidence is needed.

\textbf{Zep}\cite{rasmussen2025zep}. Zep is an agent memory architecture built around a temporal knowledge graph. It integrates conversational memory into a dynamically updated graph structure to support long-horizon retrieval and temporal state tracking.

\textbf{LangMem}\footnote{https://github.com/langchain-ai/langmem}. LangMem is a practical long-term memory framework for agents in the LangChain/LangGraph ecosystem. It provides utilities for memory extraction, updating, consolidation, and retrieval in persistent agent workflows.

\textbf{A-Mem}\cite{xu2026mem}. A-Mem is an agentic memory system inspired by Zettelkasten-style note organization. It creates structured memory notes and dynamically links related memories to support retrieval and continual updating.

\textbf{MemoryOS}\cite{kang2025memory}. MemoryOS formulates agent memory as a memory operating system with explicit storage, retrieval, and update modules. It organizes information across multiple memory levels to support persistent and personalized agent behavior.

\textbf{Mem0}\cite{chhikara2025mem0}. Mem0 is a scalable long-term memory architecture that continuously extracts, consolidates, and retrieves salient conversational information. It emphasizes production-oriented memory management for deployed AI agents.

\textbf{LightMem}\cite{fang2026lightmem}. LightMem is a lightweight memory-augmented generation framework inspired by classical cognitive memory models. It combines filtering, topic-aware organization, and offline consolidation to improve efficiency and memory quality.

\textbf{NEMORI}\cite{ma2025deserves}. NEMORI is a self-organizing memory framework motivated by cognitive principles. It adaptively distills and restructures memories to improve long-term retention and retrieval under evolving interaction histories.

\textbf{SimpleMem}\cite{liu2026simplemem}. SimpleMem is an efficient lifelong memory framework based on semantic lossless compression. It combines structured compression and adaptive retrieval to reduce memory redundancy while preserving useful information.

\textbf{HiMem}\cite{zhang2026himem}. HiMem is a hierarchical long-term memory framework that separates different forms of memory and links them through a structured hierarchy. It further supports memory evolution through conflict-aware reconsolidation.

\textbf{O-Mem}\cite{wang2025mem}. O-Mem is an omni-memory system designed for personalized, long-horizon, self-evolving agents. It emphasizes user-centric memory modeling, dynamic profile updating, and broad memory coverage across agent interactions.

\textbf{SwiftMem}\cite{tian2026swiftmem}. SwiftMem is a query-aware memory system designed to improve retrieval efficiency as memory grows. It introduces optimized indexing and retrieval mechanisms for fast long-horizon agent memory access.

\section{Detailed System Prompts}
This appendix reports representative prompt templates used in \textsc{MemIR}. Template variables such as \texttt{\{\{handle\_usual\_max\}\}} are kept in their original form for readability.

\subsection{Handle Extraction Prompt}
\begin{PromptBlock}
Extract mention handles from this page.

Use the system rules to select a small set of exact mentions that identify concrete things in this page.

Before emitting a handle:
1. Decide whether the exact mention is worth keeping.
2. Confirm it points to a specific local item in this page, not only a broad topic, domain, category, identity group, support relation, or life area.
3. Confirm it would still be a useful name if shown alone in search results.
4. Confirm it is noun-like rather than a clause or action snippet.
5. Choose the shortest exact substring that preserves that mention and distinguishes the thing.
6. Omit it if the mention is mainly a theme, value, feeling, identity expression, support phrase, or weak pointer, even when it is an exact substring.
7. Omit it if it is a clause-like snippet, verb phrase, subject-verb snippet, speaking fragment, sentence fragment, or depends on words like this, that, these, those, my, your, or a generic the-phrase to identify the item.
8. Omit it if shortening would turn it into a broad category, pronoun-like phrase, or bare generic noun.
9. Do not add weak or generic handles just to reach the usual quantity.

Output fields:
- surface_text
- support_span_ids

Field rules:
- `surface_text` must be copied exactly from one cited support span.
- `support_span_ids` must contain the span ids that ground the handle.

Output shape:
Return only a JSON object: {"handles": [...]}.

Quantity:
- Empty output is allowed.
- Usually return 1-{{handle_usual_max}} handles.
- Return {{handle_max}} only when all {{handle_max_word}} are strong, distinct, and stable.
\end{PromptBlock}

\subsection{Pivot Extraction Prompt}
\begin{PromptBlock}
You select pivot candidates from one page.

Input:
- page text
- a deterministic list of sentence-level candidate spans

Task:
A pivot marks one concrete external page item recorded by a candidate span. The candidate's main job should be to record an external event, arrangement, plan, visit, creation, application, attendance, completion, or change.

Selection steps:
1. Read the page and candidate list.
2. Judge each candidate span before thinking about its label.
3. Decide the candidate span's main meaning: does it record the external page item itself, or does it mainly explain meaning, feeling, value, identity expression, support effect, or why it matters?
4. Check that the accepted item is stated in the candidate span itself, not inferred from a concrete noun inside an interpretation sentence.
5. Keep candidates whose main job is to record a concrete external event, plan, object use, visit, creation, change, application, attendance, completion, or arrangement.
6. Skip candidates that mainly express interpretation, reaction, aspiration, value, support effect, or identity expression.
7. Group candidates that point to the same page item.
8. Pick one best candidate for each page item.
9. Emit pivots only for the selected candidates.

Do not rescue a weak candidate by writing a more concrete referent_label. If the candidate span is mainly interpretation, reaction, aspiration, or value, skip it even when it contains a concrete noun.
Do not rescue a meaning-focused sentence by turning a concrete noun or action inside it into the referent_label.

Output fields:
- candidate_ref
- support_text
- referent_label

Field rules:
- `candidate_ref` must be one of the provided candidate ids.
- `support_text` must be copied exactly from that candidate span.
- `support_text` should keep enough words to identify the concrete page item.
- `referent_label` must be short, close to page wording, and specific enough to identify the page item.

Output:
Return only a JSON object: {"pivots": [...]}.

Each pivot object includes:
- candidate_ref: the id of the accepted candidate
- support_text: exact substring copied from the accepted candidate span
- referent_label: short label for the concrete page item

Quantity:
- A page may have zero, one, or several pivots.
- Emit at most one pivot for the same concrete page item.
- Usually emit 1-{{pivot_usual_max}} pivots.
- Emit {{pivot_max}} only when the page clearly discusses {{pivot_max_word}} separate concrete page items.
\end{PromptBlock}

\subsection{Claim Writing Prompt}
\begin{PromptBlock}
# Page ClaimUnit Writer

You write page-level Claims for long-term memory.

A Claim is a locally supported fact that a later answer model can retrieve and use as evidence. It is not a page summary, not a transcript log, not one sentence per span, and not a cleaned-up paraphrase of the page. Prefer concrete facts that remain useful when retrieved beside similar memories about the same subject, object, event, work, place, image, or time.

Return 0 to `page.max_claim_units` units.

## Payload Roles
- `ordered_turn_cards`: page-local evidence cards in conversation order. Each card keeps the original order of dialogue and image items in `sequence`.
- `starter_ids`: current-page dialogue span ids that are eligible as AU anchors. Every final unit must cite at least one of these ids.
- `sequence` dialogue items: the actual text to judge for memory content. A card can produce zero, one, or many units; do not write one unit per item mechanically.
- `sequence` image items: visual evidence attached to nearby dialogue in the same turn. They can complete a fact but cannot be the only anchor.
- `boundary_context`: small previous/next-page dialogue for page-edge continuation and reference resolution only.
- `turn_handles`: local referent hints for the card. Use them to name people, objects, works, images, or other discussed items when the evidence supports the referent.
- `pivot_hints`: local continuity hints on a dialogue item. Use them to keep similar ongoing subjects apart; do not treat them as evidence by themselves.
- `time_cues`: local time hints on a dialogue item. Use them when time distinguishes an event or plan on the long-term timeline.

## Three-Step Protocol

### Step 1: Read Ordered Evidence Cards
Read each `ordered_turn_cards` card as a small evidence scene, not as independent fields.

For every dialogue item in `sequence`, decide whether it contributes durable memory content after considering adjacent sequence items, images in the same card, handles, pivot hints, time cues, and page-edge boundary context.

Write or merge content when it contributes any concrete part of a durable memory fact:
- event, change, plan, visit, win, loss, creation, recommendation, message, visual evidence, activity, or result
- state, relation, preference, evaluation, ability, habit, ownership, or constraint
- concrete value: person, object, title, count, place, visible detail, reason, result, step, ingredient, time, or target
- continuation, short answer, or confirmation that gives a value or completes a nearby fact
- pointer such as `this`, `that`, `it`, `look`, `the photo`, or `check it out` when the same card or boundary context identifies the referent or visible content

Ignore a dialogue item only when it remains pure social flow, filler, acknowledgement, generic reaction, or request wording with no concrete fact after local context is considered.

Do not make the speech act itself the remembered fact. Avoid units whose main relation is `said`, `asked`, `shared`, `showed`, `sent`, or similar conversation-act wording. Write the underlying fact instead. If no underlying fact remains, skip it.

Do not skip concrete content just because it is small, similar to another fact, or not obviously useful for the current benchmark questions. Missing a durable, concrete fact is usually worse than keeping a nearby relevant fact.

### Step 2: Compose Consumable Facts
For each memory point, write the smallest standalone fact that preserves the useful concrete value.

First find the distinguishing slot: the supported value that separates this fact from similar memories. Common slots include:
- exact name or title
- person, object, work, event, place, or visual subject
- color, visible object, visible text, count, or scene detail
- count, ordinal, size, amount, duration, or number of times
- date, relative timeline locator, or event order
- reason, result, target, condition, step, ingredient, preference, or ownership

Build each unit around:

`subject + factual relation + concrete value/detail + needed qualifier`

Use adjacent sequence items, image items, handles, pivot hints, time cues, and boundary context only to make that fact complete and unambiguous:
- resolve pronouns, generic object labels, local pointers, and vague references to visible or discussed items
- include exact titles, names, counts, colors, places, visible objects, ingredients, plans, results, and reasons when present
- include time when it distinguishes repeated events, plans, visits, wins, losses, photos, recipes, messages, projects, or scripts
- include image-visible content when a current dialogue item is a carrier for that image

Image handling:
- The image is a visual supplement to nearby same-card dialogue.
- If the dialogue points to visual evidence, write the visible objects, text, count, color, spatial relation, or other visible value that makes the fact consumable.
- Do not write a generic sharing/showing fact when the visible content is the durable memory.
- Include the image span id in support only when the unit uses visible image content, and pair it with a current-page dialogue support span.

Time handling:
- Preserve natural relative time when it locates an event, such as `last Friday`, `yesterday`, `recently`, `this weekend`, or `next month`.
- Do not treat the utterance date as a normal value to add. Use a payload date only when the remembered fact itself needs an absolute date to distinguish it and the payload clearly supports that date.
- Do not copy clock times unless the clock time itself is the remembered value.
- Do not force dates onto ordinary current-state, praise, emotion, or conversation-flow facts.

### Step 3: Merge By Answer Slot
Merge nearby evidence only when the pieces are parts of the same answer slot: same subject, same relation, and complementary details such as object plus time, event plus result, image carrier plus visible content, item plus reason, or recipe plus ingredients.

Keep separate units when facts differ by person, object, title, date, place, stage, preference, reason, outcome, image, ownership, or answer value. Similar topic is not enough to merge.

Each final unit must:
- be one standalone sentence
- include one to three `support_span_ids`
- include at least one current-page dialogue span from `starter_ids`
- cite all spans whose content is used, including image spans when visual content is used
- be concrete, non-duplicative, and free of unresolved references, source-form wording, internal field names, and invented details

Before returning, check:
- If the unit is broad enough to match many similar memories, add the locally supported distinguishing slot.
- If the unit is only social flow or weak reaction, drop it.
- If the unit mainly says someone said/asked/shared/showed something, rewrite as the underlying fact or drop it.
- If the unit still contains vague references, resolve them from the payload or drop it.

## Return Format
Return only a JSON object:
```json
{
  "units": [
    {
      "unit_text": "Complete supported sentence.",
      "support_span_ids": ["span_id"]
    }
  ]
}
```
\end{PromptBlock}

\subsection{LLM Selector Prompt}
\begin{PromptBlock}
Select bundles for the final answerer.

Work in three steps internally. Do not output the steps.

Step 1: Identify answer components.
Break the question into the distinct components that must be known to answer it.
A component may be a requested value, a list member, one counted event, one side of a comparison, a time anchor, a reason, a state before or after a change, or an input needed for simple inference.

Step 2: Cover components.
Select a bundle as `direct` when its Answer Unit or attached grounding fills one still-needed answer component.
If several bundles fill the same component, keep the one with the most specific Answer Unit or grounding.
Select a bundle as `support` when it is needed to interpret a selected direct bundle, such as clarifying its referent, time, or local context.

Step 3: Stop by coverage, not by first match.
Do not stop after the first plausible candidate if the question needs multiple components.
Continue selecting until the answer components are covered, the selection budget is reached, or no candidate fills a remaining component.
Return an empty selected list only when no candidate fills any answer component.
Select up to {{Bundle_max}} bundles.

Rank and scores are retrieval hints only. Prefer a lower-ranked bundle that fills a required component over a higher-ranked topical neighbor.

Return only a JSON object:
{
  "selected": [
    {"bundle_id": "...", "role": "direct"}
  ]
}
\end{PromptBlock}

\end{document}